\documentclass[11pt]{article}

\usepackage[utf8]{inputenc}   
\usepackage[T1]{fontenc}
\usepackage{lmodern}          
\usepackage{geometry}         
\usepackage{microtype}        
\usepackage{aliascnt}

\usepackage{mathtools, amssymb, amsthm, bbm, bm}
\usepackage{enumitem}

\usepackage[ruled,vlined]{algorithm2e}
\usepackage[noend]{algpseudocode}

\usepackage{graphicx}
\usepackage{booktabs}
\usepackage{caption}
\usepackage{subcaption}

\usepackage{threeparttable}

\usepackage{xcolor}

\usepackage{xcolor}
\definecolor{utburnt}{HTML}{BF5700}
\definecolor{charcoal}{HTML}{333333}
\definecolor{goldenrod}{HTML}{D99E00}
\definecolor{tealblue}{HTML}{007C91}

\usepackage[backref=page,bookmarks]{hyperref}
\hypersetup{
  colorlinks=true,
  linkcolor=utburnt,
  citecolor=tealblue,
  urlcolor=tealblue
}

\usepackage[nameinlink,capitalize]{cleveref}
\usepackage[numbers,sort&compress]{natbib}

\theoremstyle{plain}

\newtheorem{theorem}{Theorem}[section]

\newaliascnt{lemma}{theorem}

\aliascntresetthe{lemma}
\crefname{lemma}{lemma}{lemmas}
\Crefname{lemma}{Lemma}{Lemmas}

\newaliascnt{corollary}{theorem}
\newtheorem{corollary}[corollary]{Corollary}
\aliascntresetthe{corollary}
\crefname{corollary}{corollary}{corollaries}
\Crefname{corollary}{Corollary}{Corollaries}

\newaliascnt{proposition}{theorem}
\newtheorem{proposition}[proposition]{Proposition}
\aliascntresetthe{proposition}
\crefname{proposition}{proposition}{propositions}
\Crefname{proposition}{Proposition}{Propositions}

\newaliascnt{fact}{theorem}

\aliascntresetthe{fact}
\crefname{fact}{fact}{facts}
\Crefname{fact}{Fact}{Facts}

\newaliascnt{conjecture}{theorem}

\aliascntresetthe{conjecture}
\crefname{conjecture}{conjecture}{conjectures}
\Crefname{conjecture}{Conjecture}{Conjectures}

\newaliascnt{assumption}{theorem}
\newtheorem{assumption}[assumption]{Assumption}
\aliascntresetthe{assumption}
\crefname{assumption}{assumption}{assumptions}
\Crefname{assumption}{Assumption}{Assumptions}

\newaliascnt{definition}{theorem}
\newtheorem{definition}[definition]{Definition}
\aliascntresetthe{definition}
\crefname{definition}{definition}{definitions}
\Crefname{definition}{Definition}{Definitions}

\newaliascnt{remark}{theorem}
\newtheorem{remark}[remark]{Remark}
\aliascntresetthe{remark}
\crefname{remark}{remark}{definitions}
\Crefname{remark}{Remark}{Definitions}

\numberwithin{equation}{section}

\def\A{\mathcal{A}}

\def\F{\mathcal{C}}
\def\D{\mathcal{D}}

\def\F{\mathcal{F}}

\def\X{\mathcal{X}}

\newcommand*{\R}{{\mathbb{R}}}

\let\eps\epsilon
\let\phi\varphi

\DeclareMathOperator*{\E}{\mathbb{E}}

\DeclareMathOperator{\unif}{Unif}

\DeclareMathOperator{\poly}{poly}

\DeclareMathOperator{\sign}{sign}

\newcommand{\opt}{\mathsf{opt}}

\newcommand{\cube}[1]{\{\pm 1\}^{#1}}

\newcommand{\ignore}[1]{} 

\newcommand*{\w}{\mathbf{w}}

\newcommand*{\x}{\mathbf{x}}

\newcommand*{\Dtrainjoint}{\bar{\D}^{\mathrm{train}}}
\newcommand*{\Dtestjoint}{\bar{\D}^{\mathrm{test}}}
\newcommand*{\Dtrainmarginal}{\D^{\mathrm{train}}}
\newcommand*{\Dtestmarginal}{\D^{\mathrm{test}}}

\newcommand{\pup}{p_{\mathrm{up}}}
\newcommand{\pdown}{p_{\mathrm{down}}}

\newcommand{\Gauss}{\mathcal{N}}

\newcommand{\degreesandwich}{\ell}

\newcommand{\T}{\mathcal{T}}

\renewcommand{\P}{\mathcal{P}}

\definecolor{burntorange}{rgb}{0.8, 0.33, 0.0}

\title{Multiaccuracy as an Error Benchmark \\ for Learning with Distribution Shift}
\author{Konstantinos Stavropoulos\thanks{Part of this work was done while the author was at The University of Texas at Austin. The author was supported in part by the NSF AI Institute for Foundations of Machine Learning (IFML) and, during the 2026–27 academic year, by the Institute for Advanced Study and an anonymous donor.} \\ Institute for Advanced Study}
\date{}

\begin{document}

\maketitle

\begin{abstract}
Agnostic learning has a well-developed theory of computational and statistical complexity. For learning with distribution shift, even the choice of an appropriate error benchmark remains challenging. Testable learning with distribution shift (TDS) was introduced to develop computational foundations for this setting: using labeled training data and unlabeled test data, the learner must either produce a predictor with low test error or detect distribution shift. However, standard agnostic TDS guarantees allow a constant-factor amplification of the optimum error even when the training and test distributions coincide. They therefore do not directly recover the guarantee of classical agnostic learning.

We introduce \emph{shift multiaccuracy}, a stronger error benchmark inspired by multiaccuracy from algorithmic fairness. This yields a TDS formulation that becomes equivalent to multiaccuracy when there is no shift. Using a recent nontrivial reduction from agnostic learning to multiaccurate prediction from prior work, we obtain the first TDS framework that formally generalizes agnostic learning through an efficient reduction preserving the training marginal. The framework retains existing algorithmic upper bounds and inherits computational lower bounds from agnostic learning. Moreover, we establish a direct connection between our model and identity testing, yielding the first statistical characterization of agnostic TDS learning, in terms of Rademacher complexity.

More broadly, we argue that \emph{shift multiaccuracy} might be of independent interest in unsupervised domain adaptation, as it leads to sharper out-of-distribution generalization guarantees, and can be minimized directly, since its definition does not involve the unobserved test labels.
\end{abstract}

\thispagestyle{empty}
\newpage
\setcounter{page}{1}
\section{Introduction}

Distribution shift is a central challenge in reliable machine learning. Classical learning theory typically studies generalization under the assumption that the learner is evaluated on fresh samples from the same distribution as the one it was trained on. However, this assumption can fail in practice: the learner may be trained using carefully curated datasets and then implemented in a population that is significantly different. In response, there is, by now, a long line of research on out-of-distribution (OOD) generalization from an information theoretic \cite{ben2006analysis,blitzer2007learning,mansour2009domadapt,ben2010theory,david2010impossibility,zhang2019bridging,zhang2020localized,redko2020survey,fang2022out,awasthi2024best,hanneke2024more,hanneke2024adaptive} as well as a computational perspective \cite{goldwasser2020beyond,klivans2023testable,chandrasekaran2024efficient,klivans2024learning,goel2024tolerant,chandrasekaran2025learning,chandrasekaran2026iterative}.

While the motivation and high-level explanation of the distribution shift problem are arguably straightforward, developing a formal definition for the problem has proven to be a highly challenging task. For example, some formulations handle cases where no labels from the evaluation population are available \cite{ben2006analysis,blitzer2007learning,mansour2009domadapt,ben2010theory,zhang2019bridging}, while others exploit available labeled data from that population \cite{awasthi2024best,hanneke2024more,hanneke2024adaptive}. Even without evaluation labels, different works provide incomparable OOD generalization results for the same setting \cite{blitzer2007learning,mansour2009domadapt}. The choice of error benchmark is therefore a central part of the problem, with consequences for both the guarantees available to the learner and the complexity of obtaining them.

We focus on the \emph{unsupervised domain adaptation} case, where one only has access to labeled samples from a training distribution and unlabeled samples from a test distribution. Recent work on \emph{testable learning with distribution shift (TDS)} has provided a framework for studying this setting computationally, by allowing the learner to reject when it cannot certify a transfer guarantee \cite{klivans2023testable,chandrasekaran2024efficient,klivans2024learning,chandrasekaran2025learning}. Despite substantial progress on algorithms and lower bounds, a basic connection had remained unresolved: previous formulations of agnostic TDS learning had not been shown to formally generalize classical agnostic learning. Moreover, no general statistical characterization of agnostic TDS learning was known.

Motivated by recent work in robust learning \cite{Blanc26,klivans2026efficient}, as well as complexity theory and algorithmic fairness literature \cite{TrevisanTV09,multicalibration2018}, we propose a new error benchmark called \emph{shift multiaccuracy}. This benchmark yields sharper guarantees for unsupervised domain adaptation and leads to the first TDS framework that formally generalizes agnostic learning through an efficient reduction preserving the training marginal. We also give the first statistical characterization of agnostic TDS learning, in terms of Rademacher complexity. The new formulation connects the computational problem to multiaccurate prediction and the statistical problem to identity testing. These connections yield the first $d^{\Omega(1/\eps)}$ statistical query lower bound for TDS learning of $d$-dimensional halfspaces up to excess error $\eps$ under Gaussian training marginals.

We first explain how shift multiaccuracy sharpens the classical domain adaptation guarantee, and then describe its consequences for the formulation and complexity of agnostic TDS learning.

\paragraph{Unsupervised Domain Adaptation}
Consider the setting where the learner has access to independent samples from some labeled training distribution $\Dtrainjoint$ over $\X\times\{\pm 1\}$, as well as independent samples from some unlabeled test distribution $\Dtestmarginal$ over $\X$.

The goal is to achieve low error on a labeled test distribution $\Dtestjoint$ whose marginal on $\X$ is $\Dtestmarginal$. Classical domain adaptation guarantees account for the relationship between the training and test labels through the \emph{optimum joint error} with respect to a concept class $\F\subseteq\{\X\to\cube{}\}$:
\begin{equation}
    \lambda(\F;\Dtrainjoint,\Dtestjoint)
    = \inf_{f\in\F}\Bigl(
        \Pr_{\Dtrainjoint}[y\neq f(\x)]
        + \Pr_{\Dtestjoint}[y\neq f(\x)]
    \Bigr).
\end{equation}
One standard form of the resulting generalization bound is the following.

\begin{theorem}[Out-of-distribution generalization, e.g., \cite{mansour2009domadapt,zhang2019bridging,zhang2020localized}]\label{theorem:ood-generalization}
    The test error of any (potentially randomized) hypothesis $h$ is bounded as follows:
    \[
        \Pr_{h,\Dtestjoint}[y\neq h(\x)]
        \le \Pr_{h,\Dtrainjoint}[y\neq h(\x)]
        + \lambda(\F;\Dtrainjoint,\Dtestjoint)
        + \mathsf{disc}_{h,\F}(\Dtrainmarginal,\Dtestmarginal),
    \]
    where the final term is the hypothesis-dependent discrepancy distance (see \Cref{definition:discrepancy}).
\end{theorem}

The above result decomposes the test error into three terms. The first is the training error of $h$, which can be controlled by using training data. The second, $\lambda$, measures the compatibility of the training and test labels through a common classifier in $\F$. Some control of this relationship is necessary for meaningful domain adaptation guarantees \cite{david2010impossibility}, but $\lambda$ itself cannot generally be estimated without test labels. The third term measures the following discrepancy between the training and test marginals.

\begin{definition}[\cite{zhang2019bridging,zhang2020localized}]\label{definition:discrepancy}
    The discrepancy distance between two distributions $\Dtrainmarginal,\Dtestmarginal$ over $\X$ with respect to a (potentially randomized) hypothesis $h$ and a class $\F$ is defined as follows:
    \[
        \mathsf{disc}_{h,\F}(\Dtrainmarginal,\Dtestmarginal)
        := \sup_{f\in\F}\Bigl(
            \Pr_{h,\Dtestmarginal}[h(\x)\neq f(\x)]
            - \Pr_{h,\Dtrainmarginal}[h(\x)\neq f(\x)]
        \Bigr).
    \]
\end{definition}

For a fixed $h$, this discrepancy depends only on the two unlabeled marginals, and its estimation has sample complexity controlled by the VC dimension of $\F$ \cite{ben2006analysis,blitzer2007learning,mansour2009domadapt}. The error decomposition given in \Cref{theorem:ood-generalization} motivates a discrepancy minimization approach for domain adaptation \cite{mansour2009domadapt,zhang2020localized}, where the learner finds a hypothesis $h$ and a reweighting $\Dtrainjoint_{\mathrm{rew}}$ of the training distribution so as to minimize the following quantity:
\[
    \Pr_{h,\Dtrainjoint_{\mathrm{rew}}}[y\neq h(\x)] + \mathsf{disc}_{h,\F}(\Dtrainmarginal_{\mathrm{rew}},\Dtestmarginal)\,.
\]
Such an approach, however, can lead to an arbitrary increase of the joint error term:
\[
    \lambda(\F;\Dtrainjoint_{\mathrm{rew}},\Dtestjoint)\,.
\]
This is particularly problematic, because the parameter $\lambda$ is not observable, so the aforementioned approach would only yield a reliable error guarantee in scenarios where one has additional control of how reweighting changes the joint error parameter $\lambda$. Here, we leave the training distribution fixed and examine whether a different training objective can yield a sharper guarantee.

We focus on the agnostic case where $\lambda$ can be arbitrary, and our goal is to give the best possible OOD error guarantee under the available information. In this case, \Cref{theorem:ood-generalization} suggests a more direct approach: we may directly use the available training data to fit a near-optimal classifier $h$, and then use the unlabeled test data to estimate the associated discrepancy distance. Concretely, our goal would be to find $h$ whose training error is upper bounded by the following quantity:
\[
    \opt(\F;\Dtrainjoint) := \inf_{f\in \F} \Pr_{\Dtrainjoint}[y\neq f(\x)]
\]
This would yield the following OOD generalization bound on the test error of $h$ according to \Cref{theorem:ood-generalization}:
\begin{equation}
    \opt(\F;\Dtrainjoint) + \lambda(\F;\Dtrainjoint,\Dtestjoint) + \mathsf{disc}_{h,\F}(\Dtrainmarginal,\Dtestmarginal)\label{equation:ood-agnostic-1}
\end{equation}

This work addresses two important shortcomings of the above error benchmark. First, while it is well-known that the term $\lambda$ is unavoidable in the worst case \cite{david2010impossibility}, it is unclear whether the same is true for the term corresponding to $\opt$. Concretely, we ask: \emph{can we provide an OOD generalization bound for unsupervised domain adaptation of the following form:}
\[
    \lambda(\F;\Dtrainjoint,\Dtestjoint) + \mathsf{disc}_{h,\F}(\Dtrainmarginal,\Dtestmarginal) \textit{ ?}
\]


\paragraph{Shift Multiaccuracy}
We answer this question in the affirmative by introducing the following benchmark, defined formally in \Cref{definition:MA-ood}:
\[
    \mathsf{SMAE}(\F,h)
    := \sup_{f\in\F}\Bigl(
        \Pr_{h,\Dtestmarginal}[h(\x)\neq f(\x)]
        - \Pr_{\Dtrainjoint}[y\neq f(\x)]
    \Bigr).
\]
A small shift multiaccuracy error means that, simultaneously for every candidate $f\in\F$, the disagreement between $h$ and $f$ on the test marginal is at most the training error of $f$, up to a small additive error. The benchmark depends only on the labeled training distribution and the unlabeled test marginal. When the two marginals coincide, it becomes the multiaccuracy error $\mathsf{MAE}(\F,h)$, in the one-sided formulation used in this work. For classes closed under negation, this agrees with the classical notion of multiaccuracy up to normalization \cite{TrevisanTV09,multicalibration2018}.

The new benchmark satisfies two inequalities that connect it to both the training objective and the target prediction error:
\begin{align*}
    \mathsf{SMAE}(\F,h)
    &\le \mathsf{MAE}(\F,h)
        + \mathsf{disc}_{h,\F}(\Dtrainmarginal,\Dtestmarginal),\\
    \Pr_{h,\Dtestjoint}[y\neq h(\x)]
    &\le \mathsf{SMAE}(\F,h)
        + \lambda(\F;\Dtrainjoint,\Dtestjoint).
\end{align*}
In particular, these inequalities suggest a training objective: find a hypothesis with small multiaccuracy error. Standard multiaccuracy algorithms can achieve this using an agnostic learner as an oracle \cite{multicalibration2018}. Combining the two inequalities then yields
\[
    \Pr_{h,\Dtestjoint}[y\neq h(\x)]
    \le \lambda(\F;\Dtrainjoint,\Dtestjoint)
        + \mathsf{disc}_{h,\F}(\Dtrainmarginal,\Dtestmarginal)
        + \eps.
\]
Thus, the additional $\opt$ term in \eqref{equation:ood-agnostic-1} can be removed without reweighting the training distribution or imposing additional assumptions on the amount of shift. This yields a benchmark that does not involve test labels, while supporting a sharper error guarantee for every labeling of the test marginal.

\paragraph{Agnostic TDS Learning}
While classical domain adaptation bounds provide useful insights, estimating the associated discrepancy distances can be computationally intractable, even for simple concept classes \cite{chandrasekaran2024efficient}. To study learning with distribution shift more directly, recent work allows the learner either to abstain on individual test points \cite{goldwasser2020beyond,kalai2021efficient,goel2024tolerant,chandrasekaran2026iterative}, or to reject altogether when it cannot certify a transfer guarantee \cite{klivans2023testable,chandrasekaran2024efficient,klivans2024learning,chandrasekaran2025learning}. These frameworks admit positive results under structural assumptions on the training marginal, while leaving the test marginal unrestricted. We focus on agnostic TDS learning, whose standard formulation is the following.

\begin{definition}[Agnostic TDS Learning \cite{klivans2023testable,chandrasekaran2024efficient}]\label{definition:tds-old}
    We say that an algorithm $\A$ is an $(\eps,\delta)$-agnostic TDS learner for $\F$ with respect to a distribution $\D$ over $\X$ if the following hold for any distribution $\Dtrainjoint$ over $\X\times\cube{}$ with $\Dtrainmarginal=\D$ and any distribution $\Dtestjoint$ over $\X\times\cube{}$. The algorithm $\A$, upon receiving a large enough set of independent labeled samples from $\Dtrainjoint$ and a large enough set of independent unlabeled samples from $\Dtestmarginal$, either accepts and outputs some (randomized) hypothesis $h$, or rejects, so that the following hold with probability at least $1-\delta$:
    \begin{enumerate}[label=\textnormal{(}\alph*\textnormal{)},leftmargin=6ex]
        \item\label{item:soundness-tds} (\textit{soundness}) If $\A$ accepts, then we have:
        \[
            \Pr_{h,\Dtestjoint}[y\neq h(\x)]
            \le \lambda(\F;\Dtrainjoint,\Dtestjoint)
                + \opt(\F;\Dtrainjoint)+\eps.
        \]
        \item\label{item:completeness-tds} (\textit{completeness})
        If $\Dtrainmarginal=\Dtestmarginal$, then the algorithm accepts.
    \end{enumerate}
\end{definition}

The works of \cite{klivans2023testable,chandrasekaran2024efficient} showed that efficient agnostic TDS learning is possible whenever the class $\F$ admits low degree sandwiching approximators with respect to the distribution $\D$ (see \Cref{definition:sandwiching}). In particular, this implies a $d^{\tilde{O}(1/\eps^2)}$-time agnostic TDS learner for the class of halfspaces\footnote{A halfspace over $\R^d$ is a function of the form $f(\x) = \sign(\w\cdot \x-\tau)$ where $\w\in\R^d$ with $\|\w\|_2=1$ and $\tau\in \R$, i.e., $f$ takes the value $1$ if $\w\cdot \x\ge \tau$, and the value $-1$ otherwise.} with respect to the standard Gaussian distribution over $\R^d$. However, the only known lower bound for the problem is $d^{\tilde{\Omega}(\log(1/\eps))}$ \cite{klivans2024learning}\footnote{This lower bound works in the statistical query model, and requires that $d\le \poly(1/\eps)$.} and arises due to a barrier that holds even in the realizable setting where $\lambda=\opt=0$. 

In contrast, it is known that agnostically learning halfspaces under the Gaussian distribution requires time $d^{\Omega(1/\eps^2)}$ in the statistical query model \cite{diakonikolas2021optimality}.\footnote{Moreover, the classical result of \cite{kalai2008agnostically} together with an approximation theoretic result from \cite{diakonikolas2010ptf} imply a matching $d^{O(1/\eps^2)}$ upper bound.} To see why this lower bound does not immediately yield a corresponding lower bound for TDS learning, we now provide the formal definition of agnostic learning.

\begin{definition}[Agnostic Learning \cite{haussler92agnostic,kearns1992toward}]\label{definition:agnostic}
    An algorithm $\A$ is an $(\eps,\delta)$-agnostic learner with respect to a distribution $\D$ over $\X$ if the following hold for any distribution $\Dtrainjoint$ over $\X\times \cube{}$ with $\Dtrainmarginal= \D$. The algorithm $\A$, upon receiving a large enough set of independent labeled samples from $\Dtrainjoint$, outputs some (randomized) hypothesis $h$ such that with probability at least $1-\delta$:
    \[
        \Pr_{h,\Dtrainjoint}[y\neq h(\x)] \le \opt(\F;\Dtrainjoint) + \eps
    \]
\end{definition}

By comparing \Cref{definition:tds-old} to \Cref{definition:agnostic}, we observe that in the limit case where $\Dtrainjoint=\Dtestjoint$, the guarantee provided by the TDS learner becomes $3\cdot\opt+\eps$. Unfortunately, the existing lower bounds for agnostic learning only work for algorithms that achieve an error of $\opt+\eps$. In fact, it is known that one can agnostically learn halfspaces under the Gaussian up to error $(1+\mu)\opt+\eps$ in fully polynomial time, for any constant $\mu>0$ \cite{daniely2015ptas}. Therefore, the standard agnostic learning lower bounds do not directly imply any lower bounds for agnostic TDS learning. This leads us to the following question: \emph{Can we give a definition for agnostic TDS learning that is stronger than \Cref{definition:tds-old}, and to which there is an efficient reduction from agnostic learning?}

We answer this question by replacing the soundness criterion of \Cref{definition:tds-old} with $\mathsf{SMAE}(\F,h)\le\eps$, while retaining the same completeness condition. In the limit case $\Dtrainmarginal=\Dtestmarginal$, the new soundness criterion implies an algorithm for finding a multiaccurate predictor on the training distribution. Recent work by \cite{casacuberta2026agnostic} showed that an efficient oracle for multiaccurate prediction implies efficient agnostic learning, through a distribution-specific, efficient reduction. As such, our new formulation is formally not-easier than agnostic learning, and inherits all known agnostic learning lower bounds. See \Cref{appendix:equivalence-of-formulations} for an equivalence between the new formulation and a version of the standard agnostic TDS model. 

Moreover, we provide a statistical characterization of our agnostic TDS learning model in terms of the Rademacher complexity. Such a characterization was known for the related problem of testable agnostic learning \cite{rubinfeld2022testing,gollakota2022moment}, but extending it to the TDS setting had remained open. Our characterization is based on a new connection between TDS learning and the $\F$-identity testing problem defined by \cite{bun2026testing}. In particular, we show that an agnostic learning oracle, along with an identity testing oracle suffice for TDS learning, and that a TDS learning oracle suffices for identity testing (\Cref{theorem:tds-equiv-id-test}). Given this result, we inherit the sample complexity characterization for $\F$-identity testing provided by \cite{bun2026testing}. Our reductions are, additionally, computationally efficient, and preserve the training marginal distribution.

\section{Preliminaries}

\paragraph{Notation} We use bold letters to denote elements of $\X$, which typically is a $d$-dimensional feature space, such as $\R^d$ or the Boolean hypercube $\cube{d}$. For a vector $\x$ in $\R^d$, we denote the value of the $i$-th coordinate as $x_i$. A randomized predictor $h$ is a distribution over functions $\X\to \cube{}$. We use $h$ as a subscript in expectations and probabilities to denote that we take the expectation with respect to the distribution of $h$ over $\cube{\X}$. For a labeled distribution $\bar\D$ over $\X\times\cube{}$, we use $\D$ to denote its marginal on $\X$. Moreover, we denote with $\D^{\otimes m}$ the distribution of sets of $m$ independent samples from $\D$. We use $\D(\x)$ to denote the density of $\D$ on the point $\x$. For some of our results, we will use the following assumption:

\begin{assumption}[Chow concentration]\label{assumption:chow-concentration}
For every $\Delta\in(0,1)$, there is $m_{\mathrm{conc}}\le \operatorname{poly}(1/\Delta)d^{O(\ell)}$ such that, for every fixed joint distribution of $(\x,z)$ with $\x\sim \D$ and $|z|\le 1$,
an i.i.d.\ sample $S$ of size $m\ge m_{\mathrm{conc}}$ satisfies, with probability at least $9/10$,
\[ \max_{{\bm{\alpha}}\in\mathbb Z_{\ge 0}^d:\|{\bm{\alpha}}\|_1\le\ell}\left|\mathbb E_S[z\x^{\bm{\alpha}}]-\mathbb E[z\x^{\bm{\alpha}}]\right|\le\Delta\,, \text{ where }\x^{\bm{\alpha}} = \prod_{i\in[d]}{x_i^{\alpha_i}} \]
\end{assumption}

\paragraph{Statistical Query Algorithms} The statistical query (SQ) model was defined by \cite{kearns1998efficient}. It has served as a standard framework for computational lower bounds in learning theory. Here, we will use the following extension of the SQ model to the problem of learning with distribution shift.

\begin{definition}[Statistical Query Model]\label{definition:sq}
    Let $\tau>0$, let $\Dtrainjoint$ be a distribution over $\X\times \cube{}$, and $\Dtestmarginal$ a distribution over $\X$. We say that an algorithm $\A$ is a statistical query algorithm (SQ algorithm) with tolerance $\tau$ if $\A$ only has access to $\Dtrainjoint, \Dtestmarginal$ through adaptive queries of the form $q_{\mathrm{train}}:\X\times\cube{}\to[-1,1]$ and $q_{\mathrm{test}}:\X\to [-1,1]$, for each of which it receives values $v_{\mathrm{train}},v_{\mathrm{test}}\in\R$ with $|v_{\mathrm{train}}-\E_{\Dtrainjoint}[q_{\mathrm{train}}(\x,y)]| \le \tau$ and $|v_{\mathrm{test}}-\E_{\Dtestmarginal}[q_{\mathrm{test}}(\x)]| \le \tau$.\footnote{Throughout, complexity in the SQ model accounts for both the number of queries and the inverse of the smallest query tolerance used. Our SQ lower bounds hold in the parameter regimes of the cited results. In particular, the Gaussian lower bounds assume $\epsilon \ge d^{-c}$ for a sufficiently small constant $c>0$.}
\end{definition}

The SQ framework captures a wide range of commonly used algorithmic techniques, including gradient descent and moment methods \cite{feldman2017statistical,feldman2017statisticalquery}.

\section{Our Framework}

Motivated by recent work in robust learning \cite{Blanc26,klivans2026efficient}, as well as complexity theory and algorithmic fairness literature \cite{TrevisanTV09,multicalibration2018}, we give the following benchmark for learning with distribution shift.

\begin{definition}[Shift Multiaccuracy Error]
\label{definition:MA-ood}
    We define the shift multiaccuracy error of a (potentially randomized) hypothesis $h$ with respect to a concept class $\F$, a labeled training distribution $\Dtrainjoint$, and an unlabeled
    test distribution $\Dtestmarginal$ as follows:
    \[
        \mathsf{SMAE}(\F,h) := \sup_{f\in \F}\Bigr(\Pr_{h,\Dtestmarginal}[h(\x)\neq f(\x)] -\Pr_{\Dtrainjoint}[y\neq f(\x)]\Bigr).
    \]
    We also define the multiaccuracy error of $h$ with respect to $\Dtrainjoint$, $\mathsf{MAE}(\F,h)$, by substituting $\Dtestmarginal$ with $\Dtrainmarginal$ in the above expression.\footnote{The standard definition of multiaccuracy error is expressed in terms of the difference between the correlations of the pairs $(h,f)$, and $(y,f)$, and is two-sided. Here, we use an one-sided formulation, which is equivalent to the standard definition as long as $\F$ is closed under negation.}
\end{definition}

The notion of multiaccuracy was defined by \cite{TrevisanTV09} in the context of complexity theory. \cite{Blanc26} showed that multiaccuracy and other related notions from fairness can be used to obtain sharper error rates for robust learning. Here, we define the notion of shift multiaccuracy, which generalizes the classical multiaccuracy, and yields improved error rates for learning with distribution shift. We emphasize that there are several differences between robust learning, and learning with distribution shift. Most notably, in robust learning, the amount of noise is restricted to affect only a bounded fraction of the data, while here we make no assumptions on the amount of distribution shift. Related work has also used multiaccuracy and multicalibration to study robustness under distribution shift \cite{kim2022universal,hu2024multigroup,wu2024bridging}, but these results address different objectives or rely on additional assumptions and do not directly yield guarantees for the general unsupervised domain adaptation setting considered here.

\subsection{Domain Adaptation}

We argue that achieving a small shift multiaccuracy error is a much more reasonable benchmark in the context of learning with distribution shift. First, the shift multiaccuracy error can be expressed in terms of the information that is available to the learner, without the need to introduce unobserved test labels. Instead, this new benchmark ensures that the hypothesis $h$ would be simultaneously competitive with any candidate classifier in the class $\F$. At first glance, this might seem too much to hope for. However, the following proposition demonstrates that the shift multiaccuracy error can be bounded in terms of the multiaccuracy error of $h$ on the training distribution plus the discrepancy distance.

\begin{proposition}\label{proposition:smae-via-mae}
    $\mathsf{SMAE}(\F,h) \le \mathsf{MAE}(\F,h) + \mathsf{disc}_{h,\F}(\Dtrainmarginal,\Dtestmarginal)$, for any $h$.
\end{proposition}

\begin{proof}
    Fix any $f\in \F$. We have that:
    \[
        \Pr_{h,\Dtestmarginal}[h(\x)\neq f(\x)] \le \Pr_{h,\Dtrainmarginal}[h(\x)\neq f(\x)] + \mathsf{disc}_{h,\F}(\Dtrainmarginal,\Dtestmarginal)\,,
    \]
    due to \Cref{definition:discrepancy}. Moreover, we have:
    \[
        \Pr_{h,\Dtrainmarginal}[h(\x)\neq f(\x)] \le \Pr_{h,\Dtrainjoint}[y\neq f(\x)] + \mathsf{MAE}(\F,h)\,,
    \]
    due to \Cref{definition:MA-ood}. By combining the above inequalities, we obtain the desired result.
\end{proof}

It is well known that a randomized hypothesis $h$ with low multiaccuracy error can always be constructed from training data, as long as one has access to a weak agnostic learner for the corresponding concept class $\F$ \cite{multicalibration2018}. 

\begin{theorem}[\cite{multicalibration2018}]\label{theorem:multiaccuracy-via-agnostic}
    Let $\D$ be some distribution over $\X$. If there is an $(\eps,0.01)$-agnostic learner for $\F$ under $\D$ that runs in time $T$ and has sample complexity $m$ with $m\ge1$, then there is an algorithm that, upon receiving a large enough number of i.i.d. samples from any labeled distribution $\Dtrainjoint$ with $\Dtrainmarginal= \D$, outputs with probability at least $1-\delta$ a randomized hypothesis $h$ with $\mathsf{MAE}(\F,h)\le O(\eps)$ in time $\poly(T,1/\eps)\cdot \log(\frac{1}{\delta})$ and with sample complexity $m\cdot \tilde{O}(1/\eps^2)\cdot \log(\frac{1}{\delta})$.
\end{theorem}

Therefore, the bound on shift multiaccuracy error can be made arbitrarily close to the discrepancy between the training and test distributions, by achieving low multiaccuracy error on the training dataset.

Furthermore, in the following proposition, we show that shift multiaccuracy provides a sharp bound on the test error in the classical domain adaptation setting.

\begin{proposition}\label{proposition:ood-via-smae}
    $\Pr_{h,\Dtestjoint}[y\neq h(\x)] \le \mathsf{SMAE}(\F,h) + \lambda(\F;\Dtrainjoint,\Dtestjoint)$, for any $h$.
\end{proposition}

\begin{proof}
    Let $f^\star$ be an element of $\F$ that attains the optimum joint error $\lambda$. We have that:
    \[
        \Pr_{h,\Dtestjoint}[y\neq h(\x)] \le \Pr_{h,\Dtestjoint}[y\neq f^\star(\x)] + \Pr_{h,\Dtestjoint}[f^\star(\x)\neq h(\x)]
    \]
    Moreover, by the definition of shift multiaccuracy (\Cref{definition:MA-ood}), we have:
    \[
        \Pr_{h,\Dtestjoint}[f^\star(\x)\neq h(\x)] \le \Pr_{h,\Dtrainjoint}[f^\star(\x)\neq y] + \mathsf{SMAE}(\F,h)
    \]
    The result follows by combining the above two inequalities with the following observation:
    \[
       \lambda(\F;\Dtrainjoint,\Dtestjoint) = \Pr_{h,\Dtestjoint}[y\neq f^\star(\x)] + \Pr_{h,\Dtrainjoint}[f^\star(\x)\neq y] \qedhere
    \]
\end{proof}

Overall, we obtain the following OOD generalization bound for unsupervised domain adaptation, by combining \Cref{proposition:smae-via-mae,proposition:ood-via-smae}.

\begin{equation}
    \Pr_{h,\Dtestjoint}[y\neq h(\x)] \le \mathsf{MAE}(\F,h) + \lambda(\F;\Dtrainjoint,\Dtestjoint) + \mathsf{disc}_{h,\F}(\Dtrainmarginal,\Dtestmarginal) \label{equation:ood-new}
\end{equation}

The useful observation here is that \Cref{theorem:multiaccuracy-via-agnostic} can be used during training in order to find $h$ whose multiaccuracy error is negligible. Therefore, our formulation suggests an algorithmic approach that yields a sharper bound for OOD generalization compared to the classical result in \eqref{equation:ood-agnostic-1}.

\subsection{Agnostic TDS Learning}

Our second application concerns the TDS learning framework of \cite{klivans2023testable}. We give the following alternative definition for agnostic TDS learning.

\begin{definition}[Agnostic TDS Learning, redefined]\label{definition:tds-new}
    We say that an algorithm $\A$ is an $(\eps,\delta)$-agnostic TDS learner with respect to a distribution $\D$ over $\X$ if the following hold for any distribution $\Dtrainjoint$ over $\X\times \cube{}$ with $\Dtrainmarginal= \D$ and any distribution $\Dtestmarginal$ over $\X$. The algorithm $\A$, upon receiving a large enough set of independent labeled samples from $\Dtrainjoint$ and a large enough set of independent unlabeled samples from $\Dtestmarginal$, either accepts and outputs some (randomized) hypothesis $h$, or rejects, so that the following hold with probability at least $1-\delta$:\footnote{The only caveat of the new definition is that allowing the learner to output a randomized hypothesis might be unavoidable in the general case, while the standard agnostic TDS definition is usually achievable with deterministic output hypotheses. However, derandomizing the output hypothesis is always possible through standard, efficient techniques, as long as the target distribution $\D$ does not have heavy atoms (see, e.g., \cite{casacuberta2026agnostic}).}
    \begin{enumerate}[label=\textnormal{(}\alph*\textnormal{)},leftmargin=6ex]
    \item\label{item:soundness-tds-new} (\textit{soundness}) If $\A$ accepts, then we have:
    \[
        \mathsf{SMAE}(\F,h):= \sup_{f\in \F}\Bigr(\Pr_{h,\Dtestmarginal}[h(\x)\neq f(\x)] -\Pr_{\Dtrainjoint}[y\neq f(\x)]\Bigr) \le \eps
    \]
    \item\label{item:completeness-tds-new} (\textit{completeness}) 
    If $\Dtrainmarginal=\Dtestmarginal$, then the algorithm accepts.
    \end{enumerate}
\end{definition}

\begin{remark}
    The agnostic TDS learning formulation of \Cref{definition:tds-new} is formally stronger than the formulation of \Cref{definition:tds-old}. In particular, by combining \Cref{definition:tds-new} with \Cref{proposition:ood-via-smae}, we obtain that accuracy of an agnostic TDS learner in the soundness case is at most:
    \[
        \lambda(\F; \Dtrainjoint, \Dtestjoint) + \eps\,,
    \]
    which is a strict improvement over the soundness criterion of \Cref{definition:tds-old}.
\end{remark}

We show that all known computational complexity upper bounds for the formulation of \Cref{definition:tds-old} can also be achieved under \Cref{definition:tds-new}. To this end, we use the framework of \cite{chandrasekaran2024efficient} which separates the training and testing phases, and substitute their training phase with the multiaccuracy algorithm of \Cref{theorem:multiaccuracy-via-agnostic}, while using the same Chow-matching approach for the testing phase. Moreover, by exploiting a recent result by \cite{casacuberta2026agnostic} showing that agnostic learning is computationally equivalent to multiaccuracy, we inherit all known lower bounds for agnostic learning in the setting of \Cref{definition:tds-new}. Taken together, these results imply the following lower bound for TDS learning halfspaces:

\begin{theorem}
    The time complexity of agnostic $(\eps,0.01)$-TDS learning of halfspaces with respect to the standard Gaussian distribution over $\R^d$ in the SQ framework (\Cref{definition:sq}) is $d^{\Omega(1/\eps)}$.
\end{theorem}

\paragraph{Computational Complexity Upper Bounds} All known upper bounds for agnostic TDS learning as defined in \Cref{definition:tds-old} are based on the notion of sandwiching polynomials, which originates in the literature of pseudorandomness \cite{bazzi2009polylogarithmic,razborov2009simple,braverman2010polylogarithmic,hatamihoza}, but has recently found a number of interesting applications on the computational complexity of a wide range of primitives in reliable learning theory \cite{klivans2013moment,rubinfeld2022testing,gollakota2022moment,klivans2023testable,chandrasekaran2024efficient,goel2024tolerant,pmlr-v291-klivans25a,klivans2025the,chandrasekaran2026iterative}. We provide the definition of sandwiching polynomials below.

\begin{definition}[Sandwiching Polynomials \cite{bazzi2009polylogarithmic}]\label{definition:sandwiching}
    We say that the $\eps$-sandwiching degree of a concept class $\F$ with respect to a distribution $\D$ over $\X$ is $\ell$ if for any $f\in \F$ there exist polynomials $\pup,\pdown$ of degree at most $\ell$ such that the following hold:
    \begin{enumerate}
        \item $\pdown(\x)\le f(\x) \le \pup(\x)$ for all $\x\in\X$.
        \item $\E_{\D}[\pup(\x)-\pdown(\x)] \le \eps$.
    \end{enumerate}
    We say that the associated coefficient bound is $B$ if the sum of the absolute values of the coefficients of both $\pup$ and $\pdown$ is at most $B$.
\end{definition}

In the work of \cite{chandrasekaran2024efficient}, it was shown that $(O(\eps),\delta)$-agnostic TDS learning under \Cref{definition:tds-old} can be achieved in time $\poly(B/\eps)\cdot d^{O(\ell)}\cdot \log(1/\delta)$, whenever the $\eps$-sandwiching degree of $\F$ under $\D$ is at most $\ell$ with coefficient bound $B$. In the following theorem, we extend this result to the setting of \Cref{definition:tds-new}.

\begin{theorem}\label{theorem:upper-tds-new}
    Suppose that the $\eps$-sandwiching degree of $\F$ under $\D$ is at most $\ell$ with coefficient bound $B$ and that $\D$ satisfies \Cref{assumption:chow-concentration}. Then, there is an $(O(\eps),\delta)$-agnostic TDS learner for $\F$ with respect to $\D$ under \Cref{definition:tds-new} that runs in time $\poly(B/\eps)\cdot d^{O(\ell)}\cdot \log(1/\delta)$.
\end{theorem}

To obtain concrete computational complexity bounds for several concept classes of interest, we may combine \Cref{theorem:upper-tds-new} with existing sandwiching degree bound from the rich associated literature \cite{bazzi2009polylogarithmic,razborov2009simple,braverman2010polylogarithmic,diakonikolas2010bounded,gopalan2010fooling,diakonikolas2010ptf,kane2011kindependent,klivans2013moment,harsha2019polynomial,hatamihoza,pmlr-v336-klivans26a}. For the particular case of halfspaces under the Gaussian distribution, the sandwiching degree bound is known to be at most $\tilde{O}(1/\eps^2)$, due to \cite{diakonikolas2010bounded}. Other classes that are known to have low sandwiching degree under the Gaussian or the uniform distribution over the Boolean hypercube include intersections and arbitrary functions of halfspaces \cite{gopalan2010fooling,pmlr-v336-klivans26a}, polynomial threshold functions (PTFs) \cite{diakonikolas2010ptf,kane2011kindependent}, constant-depth circuits \cite{bazzi2009polylogarithmic,razborov2009simple,braverman2010polylogarithmic,harsha2019polynomial}, and geometric concepts with low intrinsic dimension \cite{pmlr-v336-klivans26a}. See also \Cref{table:complexity-bounds}.

In order to show \Cref{theorem:upper-tds-new}, we revisit the approach of \cite{chandrasekaran2024efficient}. In particular, their algorithm runs in two phases. First, they use labeled training data to obtain a hypothesis $h$ with near-optimal error on the training distribution. Then, they use a Chow-matching tester that estimates the so-called Chow parameters of $h$ \cite{o2008chow}, i.e., the quantities of the form:
\[
    \E_{\Dtestmarginal}[h(\x)\x^{{\bm{\alpha}}}]\,, \text{ where }\x^{\bm{\alpha}} = \prod_{i\in[d]}{x_i^{\alpha_i}}\,, \text{ and }{\bm{\alpha}}\in\{0,1,2,\dots\ell\}^d, \|{\bm{\alpha}}\|_1\le \ell\,,
\]
and rejects if at least one of the estimated quantities is more than $\Delta$-far from the corresponding quantity under the training marginal, for some small enough $\Delta= \Delta(\eps,B)$.

For the training phase, they use the following classical result on agnostic learning:

\begin{theorem}[\cite{kalai2008agnostically}]
 \label{thm:kkms}
 Let $\D$ be a distribution over $\X$ and suppose that for every $f\in\F$, there exists some polynomial $p$ of degree at most $\ell$ such that $\E_{\D}[|f(\x)-p(\x)|]\leq \epsilon$. Then there exists an $(\eps,\delta)$-agnostic learner for $\F$ with respect to $\D$ that has time and sample complexity at most $\poly(d^\degreesandwich,1/\epsilon)\cdot \log(1/\delta)$.
\end{theorem}

For the testing phase, they provide the following result:

\begin{theorem}[Chow matching tester, Theorem C.3 in \cite{chandrasekaran2024efficient}]\label{theorem:chow-mathing-tester}
    Suppose that the $\eps$-sandwiching degree of $\F$ under $\Dtrainmarginal$ is at most $\ell$ with coefficient bound $B$, and that $\Dtrainmarginal$ satisfies \Cref{assumption:chow-concentration}. Then, for any (randomized) hypothesis $h$, there is an algorithm $\T$ that has sample access to $\Dtrainmarginal$ and $\Dtestmarginal$, and query access to $h$, runs in time $\poly(d^\degreesandwich,B/\epsilon)\cdot \log(1/\delta)$ and either accepts or rejects while satisfying the following with probability at least $1-\delta$.
    \begin{enumerate}[label=\textnormal{(}\alph*\textnormal{)},leftmargin=6ex]
    \item\label{item:soundness-chow-matching} (\textit{soundness}) If $\T$ accepts, then we have:
    \[
        \mathsf{disc}_{h,\F}(\Dtrainmarginal,\Dtestmarginal) \le O(\eps)
    \]
    \item\label{item:completeness-chow-matching} (\textit{completeness}) 
    If $\Dtrainmarginal=\Dtestmarginal$, then the algorithm accepts.
    \end{enumerate}
\end{theorem}

To obtain \Cref{theorem:upper-tds-new}, we substitute the training phase with the multiaccuracy algorithm of \cite{multicalibration2018} (\Cref{theorem:multiaccuracy-via-agnostic}). We are now ready to complete the proof.

\begin{proof}[Proof of \Cref{theorem:upper-tds-new}]
    Observe that \Cref{thm:kkms}, combined with the existence of degree-$\ell$ sandwiching polynomials for $\F$ under $\D$ implies that agnostic learning for $\F$ under $\D$ can be implemented in time $T=\poly(d^\degreesandwich,1/\epsilon)\cdot \log(1/\delta)$.
    Due to \Cref{theorem:multiaccuracy-via-agnostic}, this implies that we can find a randomized hypothesis $h$ with $\mathsf{MAE}(\F,h)\le O(\eps)$ in time $\poly(d^\degreesandwich,1/\epsilon)\cdot \log(1/\delta)$. As a consequence, we obtain the following bound on shift multiaccuracy error, by using \Cref{proposition:smae-via-mae}:
    \[
        \mathsf{SMAE}(\F,h) \le \mathsf{disc}_{h,\F}(\Dtrainmarginal,\Dtestmarginal) + O(\eps)
    \]
    For the testing phase, we use the tester from \Cref{theorem:chow-mathing-tester}. If the tester rejects, then we have $\Dtrainmarginal\neq\Dtestmarginal$ (w.h.p.), so we may reject without violating the completeness criterion of \Cref{definition:tds-new}. If the tester accepts, then we have:
    \[
        \mathsf{SMAE}(\F,h) \le O(\eps)\,,
    \]
    and the randomized hypothesis $h$ satisfies the soundness condition.
\end{proof}

\paragraph{Computational Complexity Lower Bounds}

The following result from recent prior work demonstrates that achieving multiaccuracy is algorithmically equivalent to agnostic learning.

\begin{theorem}[\cite{casacuberta2026agnostic}]\label{theorem:agnostic-via-multiaccuracy}
    Suppose that $\F$ is closed under negation, and let $\D$ be some distribution over $\X$. If there is an algorithm that, upon receiving a large enough number of i.i.d. samples from any labeled distribution $\Dtrainjoint$ with $\Dtrainmarginal= \D$, outputs with probability at least $0.99$ a randomized hypothesis $h$ with $\mathsf{MAE}(\F,h)\le \eps$ in time $T$, then there is an $(O(\sqrt{\eps}),\delta)$-agnostic learning algorithm for $\F$ under $\D$ that runs in time $\poly(T,1/\eps)\log(1/\delta)$.\footnote{The reduction also preserves the SQ model, with polynomial overhead in the number of queries, the inverse query tolerance, and $1/\epsilon$.}
\end{theorem}

By observing that if we take $\Dtrainmarginal=\Dtestmarginal$, we have $\mathsf{SMAE}(\F,h) = \mathsf{MAE}(\F,h)$, we obtain the following immediate corollary.

\begin{corollary}
    Agnostic TDS learning under \Cref{definition:tds-new} is at least as hard as agnostic learning. Concretely, suppose that $\F$ is closed under negation, and let $\D$ be some distribution over $\X$. If there is an $(\eps,0.01)$-agnostic TDS learner for $\F$ under $\D$ that runs in time $T$, then there is an $(O(\sqrt{\eps}),\delta)$-agnostic learning algorithm for $\F$ under $\D$ that runs in time $\poly(T,1/\eps)\log(1/\delta)$.
\end{corollary}

As a consequence, we inherit a number of SQ lower bounds for agnostic TDS learning under the Gaussian distribution \cite{diakonikolas2021optimality,hsu2022near}, as well as the uniform distribution over the Boolean hypercube \cite{dachman2014approximate}. A selection of upper and lower bounds on the complexity of agnostic TDS learning under \Cref{definition:tds-new} in the SQ framework are summarized in \Cref{table:complexity-bounds}.

\begin{table*}[h]\begin{center}
\setlength{\tabcolsep}{3pt}
\begin{tabular}{c c c c c} 
 \toprule
 \textbf{Concept Class} & \begin{tabular}{c}\textbf{Training}  \textbf{Marginal} \end{tabular} & \begin{tabular}{c}\textbf{Upper Bounds}\end{tabular} & \begin{tabular}{c}\textbf{Lower Bounds}\end{tabular} \\ \midrule
 Halfspaces & $\Gauss_d$ or $\unif_d$ & $d^{\widetilde{O}(1/\eps^2)}$ & $d^{{\Omega}(1/\eps)}$ \\ \midrule
  \begin{tabular}{c} Intersections \& Unions \\ of $k$ Halfspaces \end{tabular} & \begin{tabular}{c} $\Gauss_d$ \end{tabular} & $d^{\widetilde{O}(k^6 / \eps^2)}$ & $d^{\widetilde{\Omega}(\log(k) + 1/\eps)}$
 \\ \midrule
\begin{tabular}{c} Degree-$q$ PTFs in \\ $k$ Dimensions \end{tabular} & $\Gauss_d$ & $d^{\widetilde{O}(q^6k^5 / \eps^4)}$ & $d^{\Omega(q^2/\eps)}$
 \\ \bottomrule
\end{tabular}
\end{center}
\caption{Upper and lower bounds on the time complexity of $(\eps,0.01)$-agnostic TDS learning in the Statistical Query model.}
\label{table:complexity-bounds}
\end{table*}

\paragraph{Sample Complexity Characterization} In order to provide a sample complexity characterization of agnostic TDS learning under \Cref{definition:tds-new}, we identify a connection to the $\F$-identity testing framework of \cite{bun2026testing}. Before establishing this connection formally, we state the sample complexity characterization in terms of the Rademacher complexity.

\begin{definition}[Rademacher Complexity]\label{definition:rademacher}
    The Rademacher complexity of a concept class $\F$ under a distribution $\D$ at sample size $m$ is defined as follows:
    \[
        {\cal R}_m(\F;\D) := \E_{(\x^{(i)})_{i}\sim \D^{\otimes m}} \E_{\bm{\sigma}\sim \cube{m}} \sup_{f\in\F} \biggr| \frac{1}{m}\sum_{i\in[m]} \sigma_i f(\x^{(i)}) \biggr|
    \]
\end{definition}

We are ready to state our sample complexity characterization result.

\begin{theorem}\label{theorem:tds-sample-complexity}
    The sample complexity of agnostic TDS learning under \Cref{definition:tds-new} is polynomially characterized in terms of Rademacher complexity. In particular, the following conditions are true for any concept class $\F$ and any distribution $\D$ over $\X$.
    \begin{enumerate}
        \item Suppose that $m$ is such that ${\cal R}_m(\F;\D) \le \eps$. Then, there is an $(O(\eps),\delta)$-agnostic TDS learner for $\F$ under $\D$ with sample complexity $O(m)\cdot \poly(1/\eps)\cdot \log(1/\delta)$.
        \item Suppose that $m\ge C/\eps^2$ for some sufficiently large universal constant $C$ and $m$ is such that ${\cal R}_m(\F;\D) \ge C\eps$. Assume, also that $\F=-\F$ and $1\in\F$. Then, every $(\eps,0.01)$-agnostic TDS learner for $\F$ under $\D$ has sample complexity $\Omega(\sqrt{m})$.
    \end{enumerate}
    Moreover, the lower bound holds even when the training labels are realized by some concept in $\F$.
\end{theorem}

A similar sample complexity characterization was already known for the related problem of testable agnostic learning \cite{rubinfeld2022testing,gollakota2022moment}. However, in testable agnostic learning there is no distribution shift, and the goal of the learner is to either certify that the output hypothesis achieves nearly optimal error on the training distribution, or that the target distributional assumption on the training marginal is violated.\footnote{For example, testable agnostic learning is an immediate generalization of distribution-specific agnostic learning. In contrast, agnostic TDS learning relates to agnostic learning through a non-trivial reduction (\Cref{theorem:agnostic-via-multiaccuracy}).} Despite the differences, \Cref{theorem:tds-sample-complexity} implies a statistical characterization of agnostic TDS learning in terms of testable agnostic learning.

We also note that the sample complexity of standard distribution-specific agnostic learning is characterized in terms of the metric entropy of the concept class, which is always smaller than the corresponding Rademacher complexity bounds. In some cases, this yields sample complexity bounds for agnostic learning that are significantly smaller than those predicted by Rademacher complexity. As such, \Cref{theorem:tds-sample-complexity} implies a general separation between the sample complexity of agnostic TDS learning, and that of distribution-specific agnostic learning. Previous results for agnostic TDS learning under \Cref{definition:tds-old} already provided separations for specific cases, such as monotone Boolean functions with respect to the uniform distribution over the Boolean hypercube and convex sets with respect to the standard Gaussian \cite{klivans2023testable}, but here we give a general separation.

The proof of \Cref{theorem:tds-sample-complexity} is based on an equivalence between agnostic TDS learning and the following testing primitive called $\F$-identity testing, which was introduced by \cite{bun2026testing}.

\begin{definition}[Identity testing \cite{bun2026testing}]\label{definition:identity-testing}
    An algorithm $\T$ is an $\F$-identity $(\eps,\delta)$-tester with respect to a distribution $\D$ over $\X$ if the following hold for any distribution $\Dtestmarginal$ over $\X$. The algorithm $\T$, upon receiving a large enough set of independent samples from $\Dtestmarginal$, either accepts or rejects, and satisfies the following with probability at least $1-\delta$:
    \begin{enumerate}[label=\textnormal{(}\alph*\textnormal{)},leftmargin=6ex]
    \item\label{item:soundness-identity-testing} (\textit{soundness}) If $\T$ accepts, then we have:
    \[
        \sup_{f\in\F}\Bigr| \E_{\D}[f(\x)] - \E_{\Dtestmarginal}[f(\x)] \Bigr| \le \eps
    \]
    \item\label{item:completeness-identity-testing} (\textit{completeness}) 
    If $\D=\Dtestmarginal$, then the algorithm accepts.
    \end{enumerate}
\end{definition}

The sample complexity of $\F$-identity testing was shown to be characterized in terms of the Rademacher complexity by \cite{bun2026testing}.

\begin{theorem}[\cite{bun2026testing}]\label{theorem:identity-testing-sample-complexity}
    The following are true for any class $\F$ and any distribution $\D$.
    \begin{enumerate}
        \item Suppose that $m$ is such that ${\cal R}_m(\F;\D) \le \eps$. Then, there is an $\F$-identity $(O(\eps),\delta)$-tester with respect to $\D$ with sample complexity $m+\poly(1/\eps)\cdot \log(1/\delta)$.
        \item Suppose that $m\ge C/\eps^2$ for some sufficiently large universal constant $C$ and $m$ is such that ${\cal R}_m(\F;\D) \ge 4\eps$. Then, every $\F$-identity $(\eps,0.01)$-tester under $\D$ requires $\Omega(\sqrt{m})$ samples.
    \end{enumerate}
\end{theorem}



Our \Cref{theorem:tds-sample-complexity} follows from combining \Cref{theorem:identity-testing-sample-complexity} with the following connection between $\F$-identity testing and agnostic TDS learning, and the fact that Rademacher complexity gives an upper bound on the sample complexity of distribution-specific agnostic learning as well.

\begin{theorem}\label{theorem:tds-equiv-id-test}
    The following are true for any concept class $\F$ and any distribution $\D$ over $\X$.
    \begin{enumerate}
        \item Suppose that there is an $\F$-identity $(\eps,0.01)$-tester with respect to $\D$ that runs in time $T_{\mathrm{test}}$ and has sample complexity $m_{\mathrm{test}}$, and an $(\eps,0.01)$-agnostic learner for $\F$ under $\D$ that runs in time $T_{\mathrm{train}}$ and has sample complexity $m_{\mathrm{train}}$, where $m_{\mathrm{train}}\ge 1$. Then, there is an $(O(\eps),\delta)$-agnostic TDS learner for $\F$ with respect to $\D$ that runs in time $O(T_{\mathrm{test}}) \cdot \poly(T_{\mathrm{train}}, 1/\eps)\cdot \log(1/\delta)$ and has sample complexity $(O(m_{\mathrm{test}}) + m_{\mathrm{train}}\cdot \poly(1/\eps))\cdot  \log(1/\delta)$.
        \item Suppose that $\F$ is closed under complement, that the constant function $f_+\equiv 1$ is in $\F$, and that there is an $(\eps,0.01)$-agnostic TDS learner for $\F$ with respect to $\D$ that runs in time $T_{\mathrm{tds}}$ and has sample complexity $m_{\mathrm{tds}}$. Then, there is an $\F$-identity $(O(\eps),\delta)$-tester with respect to $\D$ that runs in time $O(T_{\mathrm{tds}}\cdot \log(1/\delta))$ and has sample complexity $O(m_{\mathrm{tds}}\cdot \log(1/\delta))$.
    \end{enumerate}
\end{theorem}

The reduction from TDS learning to agnostic learning plus identity testing uses the identity tester to certify that the localized discrepancy between $\Dtrainmarginal$ and $\Dtestmarginal$ is small with respect to a randomized hypothesis $h$ that is found using training data and has low multiaccuracy error on the training distribution. To achieve this, we make use of a result from \cite{bun2026testing} which precisely shows that $\F$-identity testing can be used for localized discrepancy testing. This is shown in Lemma 4.7 of the arXiv version of \cite{bun2026testing}. The proof involves a decomposition of the discrepancy into a number of carefully constructed terms, each of which is either a valid instance of the identity testing problem, or can be tested directly with an efficient procedure. The other direction is a straightforward reduction of identity testing to TDS learning. We give the full proof of \Cref{theorem:tds-equiv-id-test} in \Cref{appendix:proof-of-equiv}.

\begin{remark}
    The reduction from identity testing to TDS learning uses training labels that are generated by some function that lies within the class $\F$ (in particular, the constant $1$ function). Therefore, we obtain the surprising corollary that Rademacher complexity characterizes the sample complexity of TDS learning even in the realizable setting, as long as the error benchmark is shift multiaccuracy, as per \Cref{definition:tds-new}.
\end{remark}



\section*{Statement of AI Usage}

The definition of shift multiaccuracy and the formulation of the learning framework were developed by the author. AI tools were used to assist with literature search, exposition, and proof checking, as well as to identify connections and auxiliary observations, including the connection to $\F$-identity testing and the characterization in \Cref{proposition:equivalence-of-formulations}.

\section*{Acklowledgements} The author would like to thank Adam Klivans for useful comments on earlier versions of this manuscript, as well as S\'ilvia Casacuberta, Parikshit Gopalan, Varun Kanade, Omer Reingold, and Pranay Tankala for useful discussions.

\bibliographystyle{alpha}
\bibliography{refs}

\newpage
\appendix

\section{Connection to Previous Formulation}\label{appendix:equivalence-of-formulations}

We will now demonstrate that while our shift multiaccuracy framework illuminates several technical aspects of learning with distribution shift, it is actually equivalent to a sharper version of the classical formulation. In particular, we show the following result:

\begin{proposition}\label{proposition:equivalence-of-formulations}
    The following holds for any randomized hypothesis $h$, any concept class $\F$, and any distributions $\Dtrainjoint,\Dtestmarginal$.
    \[\mathsf{SMAE}(\F,h) = \sup_{\Dtestjoint\in\mathrm{L}(\Dtestmarginal)} \Bigr( \Pr_{h,\Dtestjoint}[y\neq h(\x)] - \lambda(\F;\Dtrainjoint,\Dtestjoint) \Bigr)\,,\]
    where $\mathrm{L}(\Dtestmarginal)$ is the set of labeled distributions over $\X\times{\cube{}}$ whose marginal on $\X$ is $\Dtestmarginal$.
\end{proposition}

\begin{proof}
    The lower bound is implied by \Cref{proposition:ood-via-smae}. For the upper bound, consider the case where $\Dtestjoint = (\Dtestmarginal,f)$ and where $f\in\F$. Then, we have:
    \[
        \lambda(\F;\Dtrainjoint,\Dtestjoint) \le \Pr_{\Dtrainjoint}[y\neq f(\x)] \quad \text{and} \quad \Pr_{h,\Dtestjoint}[y\neq h(\x)] = \Pr_{h,\Dtestjoint}[f(\x)\neq h(\x)]
    \]
    Therefore:
    \begin{align*}
        \sup_{\Dtestjoint\in\mathrm{L}(\Dtestmarginal)} \Bigr( \Pr_{h,\Dtestjoint}[y\neq h(\x)] - \lambda(\F;\Dtrainjoint,\Dtestjoint) \Bigr) &\ge \sup_{f\in\F} \Bigr( \Pr_{h,\Dtestjoint}[f(\x)\neq h(\x)] - \Pr_{\Dtrainjoint}[y\neq f(\x)] \Bigr) \\
        &= \mathsf{SMAE}(\F,h) \qedhere
    \end{align*}
\end{proof}

\begin{remark}
    In light of \Cref{proposition:equivalence-of-formulations}, the new agnostic TDS definition (\Cref{definition:tds-new}) is equivalent to the version of \Cref{definition:tds-old} obtained by removing the optimum training error term $\opt$ and requiring the resulting soundness inequality to hold simultaneously for all $\Dtestjoint\in\mathrm{L}(\Dtestmarginal)$, on a single event of probability at least $1-\delta$.
\end{remark}

\section{Proof of \Cref{theorem:tds-equiv-id-test}}\label{appendix:proof-of-equiv}

We will show each direction separately.

    \paragraph{Reducing agnostic TDS to agnostic learning plus identity testing}
    We begin by using the agnostic learner as an oracle for the algorithm of \cite{multicalibration2018} (see \Cref{theorem:multiaccuracy-via-agnostic}) to obtain with probability at least $1-\delta/8$ a randomized hypothesis $h$ with $\mathsf{MAE}(\F,h)\le O(\eps)$ under the training distribution $\Dtrainjoint$. Then, we use the $\F$-identity testing oracle to either detect that $\Dtrainmarginal\neq\Dtestmarginal$ or certify the following bound on the discrepancy:
    \[
        \mathsf{disc}_{h,\F}(\Dtrainmarginal,\Dtestmarginal)
        := \sup_{f\in\F}\Bigl(
            \Pr_{h,\Dtestmarginal}[h(\x)\neq f(\x)]
            -\Pr_{h,\Dtrainmarginal}[h(\x)\neq f(\x)]
        \Bigr)\le O(\eps).
    \]
    Recall that this would suffice for soundness, due to \Cref{proposition:smae-via-mae}.
    
    A useful observation is that the zero-one loss can be rewritten in terms of correlations, i.e., it is always true that
    $\Pr[h(\x)\neq f(\x)]=(1-\E[h(\x)f(\x)])/2$.
    This allows us to focus on the following sufficient condition:
    \begin{equation}
        \sup_{f\in\F}\Bigl|
            \E_{h,\Dtrainmarginal}[h(\x)f(\x)]
            -\E_{h,\Dtestmarginal}[h(\x)f(\x)]
        \Bigr|\le O(\eps).\label{equation:sufficient-condition}
    \end{equation}
    Observe that the above condition resembles the soundness criterion of $\F$-identity testing, with the only difference being the introduction of $h$ inside the compared expectations. Lemma 4.7 in \cite{bun2026testing} shows that, nevertheless, the $\F$-identity tester can be used to certify the desired bound. Below, we provide a self contained proof which is based on the proof provided by \cite{bun2026testing}. For the tests that follow, we use fresh samples.
    
    Consider a reweighting function $w:\X\to \R$ defined as follows:
    \[
        w(\x)=\frac{1+\E_h[h(\x)]}{4}\in[0,1/2].
    \]
    Recall that $\D=\Dtrainmarginal$. For ${\cal P}\in\{\D,\Dtestmarginal\}$, define
    \[
        {\cal P}_w(\x)
        =w(\x)\cdot{\cal P}(\x)
        +\frac{\E_{{\cal P}}[1-w]}{\E_{\D}[1-w]}\cdot
          (1-w(\x))\cdot \D(\x).
    \]
    To sample from ${\cal P}_w$, we draw $\x\sim{\cal P}$ and keep
    it with probability $w(\x)$. Otherwise, we repeatedly draw
    independent samples $\x'\sim\D$, accepting each with probability
    $1-w(\x')$, and output the first accepted sample. This replacement
    sample has distribution proportional to $(1-w)\D$, and requires
    at most two draws from $\D$ in expectation since $1-w\ge1/2$. A coin with success probability $w(\x)$ is implemented by independently evaluating $h(\x)$ once and declaring success precisely when $h(\x)=1$ and a fair coin lands heads. 

    Using the distribution $\P_w$, we give the following decomposition of the left-hand side of \eqref{equation:sufficient-condition}:
    \begin{align}
        \E_{h,\D}[h(\x)f(\x)]
            -\E_{h,\Dtestmarginal}[h(\x)f(\x)] &= \E_{\D}[(4w(\x)-1)f(\x)]
            -\E_{\Dtestmarginal}[(4w(\x)-1)f(\x)]\label{equation:sufficient-2}
    \end{align}
    Observe that:
    \[
        w(\x)\cdot{\cal P}(\x) = {\cal P}_w(\x) - \frac{\E_{{\cal P}}[1-w]}{\E_{\D}[1-w]}\cdot
          (1-w(\x))\cdot \D(\x)
    \]
    Therefore:
    \begin{equation}
        \E_{\P}[w(\x)f(\x)] = \E_{\P_w}[f(\x)] - \frac{\E_{\P}[1-w(\x)]}{\E_{\D}[1-w(\x)]} \E_{\D}[f(\x)(1-w(\x))]\label{equation:sufficient-3}
    \end{equation}
    We define the following three error terms:
    \begin{align*}
        \Delta_1(f) &= \E_{\D}[f(\x)] - \E_{\Dtestmarginal}[f(\x)] \\
        \Delta_2(f) &= \E_{\D_w}[f(\x)] - \E_{\Dtestmarginal_w}[f(\x)] \\
        \Delta_3(f) &= \frac{\E_{\D}[f(\x)(1-w(\x))]}{\E_{\D}[1-w(\x)]} \cdot \Bigr( \E_{\Dtestmarginal}[w(\x)] - \E_{\D}[w(\x)] \Bigr) \le \Bigr| \E_{\Dtestmarginal}[w(\x)] - \E_{\D}[w(\x)] \Bigr|
    \end{align*}

    By combining equations \eqref{equation:sufficient-2} and \eqref{equation:sufficient-3} with the definitions of $\Delta_1,\Delta_2,\Delta_3$, we obtain that:
    \[
        \sup_{f\in\F}\Bigr| \E_{h,\D}[h(\x)f(\x)]
            -\E_{h,\Dtestmarginal}[h(\x)f(\x)] \Bigr| \le 4 \sup_{f\in \F}\Bigr( |\Delta_1(f)| + |\Delta_2(f)| + |\Delta_3(f)| \Bigr) 
    \]

    Therefore, it suffices to certify that each $|\Delta_i|$ is at most $O(\eps)$. For $\Delta_1$, we may run the $\F$-identity tester on $\Dtestmarginal$ (after amplifying the success probability to $1-\delta/16$ via repetition) and reject if it rejects. 

    For $\Delta_2$, we similarly run the $\F$-identity tester on $\Dtestmarginal_w$. The crucial observation here is that $\D_w=\D$, so we may still use the same $\F$-identity tester. Substituting $\P=\D$ in the definition of $\P_w$, we indeed obtain:
    \[
        \D_w(\x)=(w(\x)+1-w(\x))\D(\x)=\D(\x)\,.
    \]

    Finally, for $\Delta_3$, we may directly estimate the expectation of the function $w$ under the distributions $\D$ and $\Dtestmarginal$ and reject if the estimated quantities are more than $\eps$ away. This yields an additional $O(1/\eps^2)\cdot \log(1/\delta)$ sample overhead.

    \paragraph{Reducing identity testing to TDS learning} For the other direction, we simply run the TDS learner on the training distribution $\Dtrainjoint= (\D,1)$, i.e., the distribution where $\x\sim \D$ and the labels are always $1$. The unlabeled test distribution is the tested distribution $\Dtestmarginal$. If the TDS learner rejects, then we know that $\Dtrainmarginal\neq\Dtestmarginal$ with probability at least $0.99$. If the TDS learner accepts, and since $\F=-\F$, then we have the following for all $f\in\F$:
    \[
        \Bigr|\Pr_{\Dtestmarginal}[h(\x)\neq f(\x)] - \Pr_{\D}[1\neq f(\x)] \Bigr| \le \eps
    \]
    In particular, since $1\in\F$, we have:
    \[
        \Pr_{\Dtestmarginal}[h(\x)\neq 1] \le \eps
    \]
    \begin{align*}
        \Bigr| \E_{\D}[f(\x)] - \E_{\Dtestmarginal}[f(\x)]\Bigr| &= 2\cdot \Bigr| \Pr_{\Dtestmarginal}[f(\x)\neq 1] - \Pr_{\D}[f(\x)\neq 1]\Bigr| \\
        &\le 2\cdot \Bigr| \Pr_{\Dtestmarginal}[f(\x)\neq 1] - \Pr_{\Dtestmarginal}[h(\x)\neq f(\x)]\Bigr| + 2\eps \\
        &\le 2\cdot \Pr_{\Dtestmarginal}[h(\x)\neq 1] + 2\eps \tag{by triangle inequality}\\
        &\le O(\eps)
    \end{align*}
    Therefore, with probability at least $0.99$, we have that if the TDS learner accepts, then the soundness criterion for identity testing is satisfied.
    We may repeat the whole procedure $O(\log(1/\delta))$ times to amplify the probability of success.

\end{document}